\documentclass[11pt]{article}

\usepackage{cvpr}
\usepackage{times}
\usepackage{epsfig}
\usepackage{graphicx}
\usepackage{amsmath}
\usepackage{amssymb}
\usepackage{amsthm}
\usepackage{placeins}
\usepackage{dsfont}

\newcommand{\R}{\mathbb{R}}
\newcommand{\PP}{\mathcal{P}}
\newcommand{\rr}{\rule[-2mm]{0mm}{6mm}}
\newcommand{\E}{\mathcal{E}}
\newcommand{\BB}{\mathcal{B}} 
\newcommand{\KK}{\mathcal{K}}
\newcommand{\suppress}[1]{} 

\DeclareMathOperator*{\argmax}{\arg\,max}

\DeclareMathOperator*{\med}{median}	
\DeclareMathOperator*{\Co}{Co}	

\usepackage[pagebackref=true,breaklinks=true,letterpaper=true,colorlinks,bookmarks=false]{hyperref}
\cvprfinalcopy 

\def\cvprPaperID{} 

\ifcvprfinal\fi
\begin{document} 

\title{Co-Sparse Textural Similarity for Image Segmentation}

\author{Claudia Nieuwenhuis $\quad$ Daniel Cremers\\
Department of Computer Science\\
Technische Universit\"at M\"unchen
85748 Garching, Germany \\
{\tt\small claudia.nieuwenhuis@in.tum.de, cremers@tum.de} \\
%
\and
Simon Hawe $\quad$ Martin Kleinsteuber\\
Department of Electrical Engineering and Information Technology\\
Technische Universit\"at M\"unchen
80333 M\"unchen, Germany \\
{\tt\small \{simon.hawe,kleinsteuber\}@tum.de}
}

\maketitle 
\begin{abstract}
We propose an algorithm for segmenting natural images based on texture and color information, which leverages the co-sparse analysis model for image segmentation within a convex multilabel optimization framework. As a key ingredient of this method, we introduce a novel textural similarity measure, which builds upon the co-sparse representation of image patches. We propose a Bayesian approach to merge textural similarity with information about color and location. Combined with recently developed convex multilabel optimization methods this leads to an efficient algorithm for both supervised and unsupervised segmentation, which is easily parallelized on graphics hardware. The approach provides competitive results in unsupervised segmentation and outperforms state-of-the-art interactive segmentation methods on the Graz Benchmark.
\end{abstract}

\section{Introduction}
The segmentation of natural images is a fundamental problem in
computer vision. It forms the basis of many high-level algorithms such
as object recognition, image annotation, semantic scene analysis,
motion estimation, and 3D object reconstruction. Despite
its importance, the task of unsupervised segmentation is highly
ill-posed and admittedly hard to evaluate since the quality of a
segmentation result depends on the subsequent application.
The hand-drawn ground truth of the Berkeley benchmark~\cite{martin_et_al01} well
demonstrates that different users have very different understandings
of the same scene:  While some people see the windows of a skyscraper as
a single texture of the building, others see them as separate objects.

\begin{figure}[tp]
\tabcolsep0.4mm
  \begin{center}
    \begin{tabular}{cc}
      \includegraphics[width=0.49\linewidth]{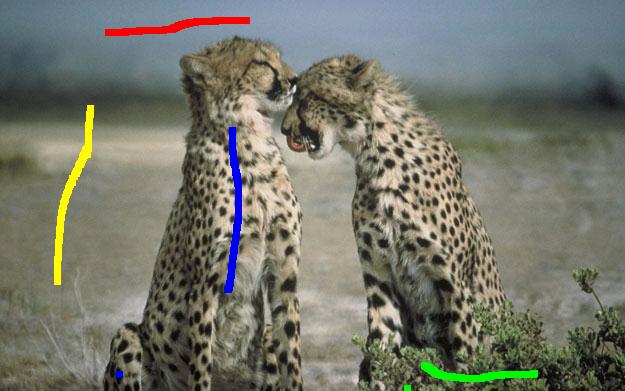}&
      \includegraphics[width=0.49\linewidth]{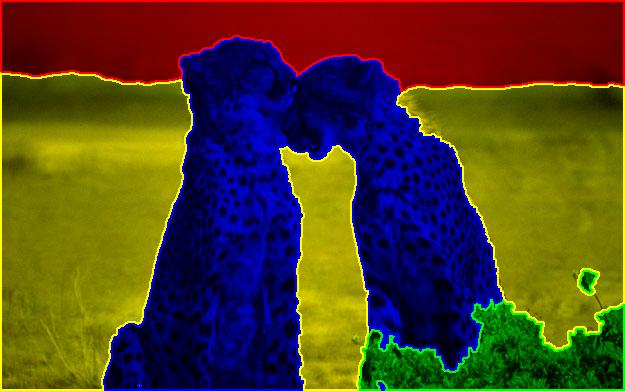} \\
       \includegraphics[width=0.49\linewidth]{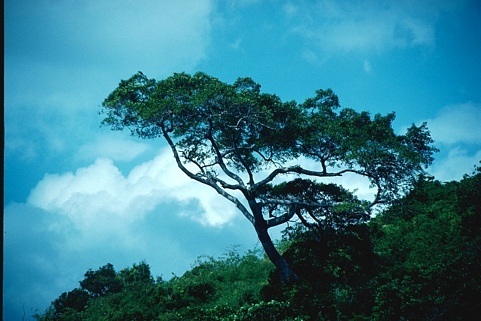} &
       \includegraphics[width=0.49\linewidth]{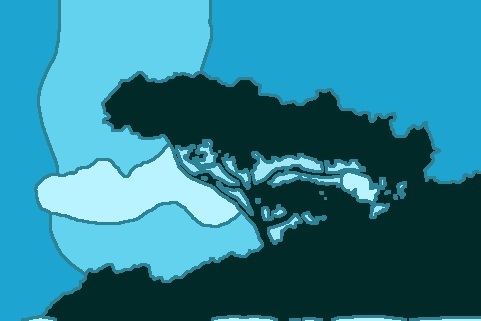} \\ 
      \end{tabular}
\end{center}
\caption{Results of the proposed supervised (top) and unsupservised
  (bottom) segmentation algorithm, which is based on the co-sparse
  analysis model to measure textural similarity and a convex variational
  multilabel method.}
\label{fig:teaser}
\vspace{-3mm}
\end{figure}

In this paper, we focus on two scenarios, namely {\em unsupervised
  segmentation} where no prior knowledge is given and hence
quantitative validation is questionable, and {\em
  supervised segmentation} where  ambiguities are solved by additional
user input (scribbles or bounding boxes) and a clearly defined ground
truth for performance evaluation is available --  see Figure
\ref{fig:teaser}. 
In both cases, one can compute data likelihoods using color
\cite{boykov-jolly01,Felzenszwalb-et-al-04,mignotte12,Rother-et-al-04,Unger-et-al-08},
texture~\cite{arbelaez_et_al11,santner10,Shi-Malik-97}, or
location~\cite{Brox-Cremers-09,nieuwenhuis_cremers_pami12}. While all
features carry relevant information, for natural images texture
features are particularly relevant. To model the dependence of texture
on image location and color we extend the approach in \cite{nieuwenhuis_cremers_pami12} to a spatially varying color \emph{and} texture model.

To extract textural information from images, methods based on sparse representations are quite successful \cite{mairal:2008}. Commonly, sparsity is exploited via the synthesis model, which assumes that every image patch can be approximated as a linear combination of a suitable dictionary and a very sparse vector or code. With this, the textural information is encoded in the set of active dictionary atoms, \ie the support of the sparse code. Finding this set, however, requires to solve a costly optimization problem.

In this paper, we propose a more efficient way to obtain textural information by employing the \emph{co-sparse analysis model} \cite{elad:07,nam:2013}. In this model, the sparse image representation is determined efficiently by a simple matrix vector product. Based on the theory developed in \cite{elad:07,nam:2013} we derive a novel textural similarity measure for image patches and demonstrate it can be introduced to image segmentation.  
Figure \ref{fig:teaser} shows that the proposed measure combined with an efficient convex multilabel approach generates convincing results for both supervised and unsupervised segmentation problems.

In particular, we make the following contributions \vspace{-0.2cm}
\begin{itemize}
\item The proposed approach combines the co-sparse analysis model, recent convex relaxation techniques, and label cost priors within a single optimization problem. \vspace{-0.3cm}
\item The co-sparse analysis model is leveraged for image segmentation through a novel texture similarity measure. \vspace{-0.3cm}
\item The method extends the purely color based approach in \cite{nieuwenhuis_cremers_pami12} by explicitly modeling the dependence between texture, color and location leading to a spatially varying color and texture model. \vspace{-0.3cm}
\item The approach does not rely on superpixels and thus avoids potentially incorrect pixel aggregations. 
\end{itemize}



\section{Co-Sparse Textural Similarity}
\label{sec:textural_similarity}


The co-sparse analysis model \cite{elad:07,nam:2013} is based on the assumption that if $\mathbf{s}\in\R^N$ denotes a vectorized image patch, there exists an analysis operator $\mathbf{O} \in \R^{k \times N}$ with $k>N$ such that $\mathbf{a}:=\mathbf{O} \mathbf{s}$ is sparse. We refer to $\mathbf{a} \in \R^k$ as the \emph{analyzed version of $\mathbf{s}$}. The two major differences to the more commonly known synthesis model are: (i) the sparse code is found via a simple matrix vector multiplication and (ii) the \emph{zero} entries of $\mathbf{a}$ are the informative coefficients describing the underlying signal. Concretely, the textural structure of $\mathbf{s}$ is encoded in its co-support
\begin{equation}
\Co(\mathbf{a}):=\{j~|~a_j=0 \},
\end{equation}
where $a_j$ denotes the $j$-th entry of $\mathbf{a}$. Geometrically, $\mathbf{s}$ is orthogonal to all rows that determine the co-support and thus lies in the intersection of the respective hyperplanes. 

Note, that the better a signal fits to the co-sparse analysis model, the sparser is its analyzed version, \ie the larger is its co-support. Since our ultimate goal is to discriminate between distinctive textures in natural images, 
a measure of textural similarity should 
better distinguish between representative patches, \ie patches that fit the co-sparse analysis model of natural image patches, while discriminating moderately for "outlier"-patches, \ie patches that seldom occur in natural images. This motivates us to measure the textural similarity between two patches via
\begin{equation}\label{eq:tsm_clean}
TSM_{\mathbf{O}}(\mathbf{s}_1,\mathbf{s}_2):=\sum_{j=1}^{k}| {\mathds{1}}_{\Co(\mathbf{O}\mathbf{s}_1)}(j) - 
{\mathds{1}}_{\Co(\mathbf{O}\mathbf{s}_2)}(j) |,
\end{equation}
where $\mathds{1}_A$ is the indicator function of a set, \ie $\mathds{1}_A(j)=1$ if $j \in A$ and zero otherwise. This measure has two desired properties: 1) it distinguishes sensibly between patches that fit the model well, \ie patches with a large co-support, 2) it does not heavily discriminate between patches that fit the model less.

In our experiments, we employ an analysis operator learned according to~\cite{Hawe2012} from patches extracted from natural images. As we only want to gather textural information independent of varying illumination conditions, we follow the simple bias and gain model and use patches that have been normalized to zero-mean and unit-norm, \ie $\sum_i s_i=0$ and $\|\mathbf{s}\|_2=1$. We exclusively consider such patches for the remainder of the paper.

To identify an "average" textural structure from a set of $m$ patches $\mathcal{S}=\{\mathbf{s}_1,\dots,\mathbf{s}_m\}$ that serves as their textural representative, we provide the following definition. Let $\mathbf{r} \in \mathbb{R}^N$ be a patch with analyzed version $\mathbf{z}=\mathbf{O}\mathbf{r}$, then we say that $\mathbf{r}$ is a \emph{textural representative of $\mathcal{S}$} if %
\begin{equation}
j \in \Co(\mathbf{z}) \; \Leftrightarrow \; \sum\limits_{i=1}^m \mathds{1}_{\Co(\mathbf{O}\mathbf{s}_i)}(j) \geq \tfrac{1}{2} m.
\end{equation}
In other words, the co-support of the analyzed version of a textural
representative is determined by the majority of the analyzed versions
of the elements in $\mathcal{S}$. Note, that this definition is
neither constructive in the sense that the representative $\mathbf{r}$
can be easily deduced given $\mathcal{S}$, nor that it is uniquely
determined. Nonetheless, it proves useful for clustering image patches
according to their texture and can be efficiently approximated.

Up to now, we considered truly co-sparse image patches, \ie patches whose analyzed versions contain many coefficients that are exactly zero. However, this is an idealized assumption and in practice those patches are not truly co-sparse but rather contain many coefficients that are close to zero.
To account for this, we introduce the mapping $\iota_\sigma\colon \mathbb{R}^{k} \to \mathbb{R}^{k}$ as a smooth approximation of the indicator function of the co-support,  
%
which is defined component-wise with a free parameter $\sigma > 0$ as
\begin{equation}\label{eq:map_tsm}
\left(\iota_\sigma(\mathbf{a})\right)_j=\exp(-a_j^2/\sigma).
\end{equation}
In fact, it is easily seen that
$\mathds{1}_{\Co(\mathbf{a})}(j) = \lim_{\sigma \to 0} (\iota_\sigma(\mathbf{a}))_j$
and
$\lim_{a_j \to 0} (\iota_\sigma(\mathbf{a}))_j=1$.

Using this approximation of the co-support, our textural similarity measure in \eqref{eq:tsm_clean} of two patches $\mathbf{s}_1$ and $\mathbf{s}_2$ associated with the analysis operator $\mathbf{O}$ and $\sigma$ is approximated by
\begin{equation}\label{eq:tsm}
TSM_{\mathbf{O},\sigma}(\mathbf{s}_1,\mathbf{s}_2)= \|\iota_\sigma(\mathbf{O}\mathbf{s}_1)-\iota_\sigma(\mathbf{O}\mathbf{s}_2)\|_1,
\end{equation}
with $\| \cdot \|_1$ denoting the $\ell_1$-norm. 
Using this, we approximate the co-support of a textural representative of a set $\mathcal{S}$ by the median of the set $\mathcal{A}=\{ \mathbf{a}_i~|~i=1,\dots,m \}$ where $\mathbf{a}_i=\iota_\sigma(\mathbf{O s}_i)$, \ie
\begin{equation} \label{eq:approx_cosupp}
{\mathds 1}_{\Co(\mathbf{z})}(j) \approx (\med (\mathcal{A}))_j.
\end{equation}
Note that this is in accordance to the well-known fact that the centroid of a cluster with respect to the $\ell_1$-distance is given as the median of all corresponding cluster points.


\section{Variational Co-sparse Image Segmentation}
\label{sec:variational}
In this section, we derive a Bayesian inference formulation for both supervised and unsupervised image segmentation based on the proposed textural similarity measure. We explicitly model the dependence of texture and color on the location in the image to account for texture variations within regions, \eg a sky which is partially covered by clouds.

\subsection{A Space Variant Texture and Color Distribution}
For an image domain $\Omega \subset \mathbb{R}^2$, let $I: \Omega \to
\mathbb{R}^d$ denote the input color (or gray scale)
image. 
The segmentation problem can be solved by computing a labeling
$l\!:\!\Omega\! \to \{1,..,n\}$ that indicates, which of the $n$ regions
each pixel belongs to, \ie $\Omega_i:=\{x\big|l(x)=i\}$. In a Bayesian framework the labeling $l$ can be computed by maximizing the conditional probability
\begin{equation}
\label{eq:bayes2}
	\arg\max_l \mathcal{P}(l\,|\,I) = \arg\max_l \mathcal{P}(I\,|\,l)\;
        \mathcal{P}(l).  
\end{equation}
In the following, we will model the dependence of color and texture on to the image location. 
Let $\mathbf{s}_x$ denote a small gray value texture patch centered at
pixel $x$. With the assumption that a pixel color depends on the local texture
$\mathbf{s}_x$ and its location $x$, but is independent of the label of
other pixels we obtain
\begin{equation} \label{eq:apost}
\mathcal{P}(I\,|\,l) = \prod_{i=1}^n \prod_{x\in\Omega} \mathcal{P}(I(x),\mathbf{s}_x,x\,|\,l(x) = i). 
\end{equation}
In the following, we derive the required probability $\mathcal{P}(I(x),\mathbf{s}_x,x\,|\,l(x) = i)$ that a pixel $x$ with texture patch $\mathbf{s}_x$ belongs to segment $i$.
\paragraph{Supervised Segmentation} 
Given the set of scribble samples consisting of location, color, and texture patches for each segment $i$, \ie
\begin{equation}
	S_i := \left\{(x_{ij},I_{ij},\mathbf{s}_{x_{ij}}), \; j=1,..,m_i\right\}
\end{equation}
we can - assuming independence - compute the likelihood of a pixel for belonging to region $i$ as the joint probability 
\begin{equation}\label{eq:joined}
\!\!\!\!\!\mathcal{P}(I(x), \mathbf{s}_x, x \,| l\! =\! i) = \mathcal{P}(I(x),x\, | l\! =\! i) \mathcal{P}(\mathbf{s}_x,x\,| l\! =\! i).
\end{equation}
Following~\cite{nieuwenhuis_cremers_pami12} we compute a space-dependent color likelihood based on the Parzen kernel density estimator~\cite{Parzen-62}
\begin{align} \label{eq:color} 
 	\mathcal{P}(I(x),x\,|\, l\! =\! i) =  \tfrac{1}{m_i} \sum_{j=1}^{m_i} k_{\rho_i}(x-x_{ij}) k_{\sigma}(I-I_{ij})
\end{align} 
where $k$ denotes a kernel function. We use a Gaussian with variance
$\sigma = 1.3$. Parzen density estimators come with the advantage that
they can represent arbitrary kinds of color distributions and provably
converge to the true density for infinitely many samples. The
estimated color distributions have shown to yield accurate results for
interactive segmentation~\cite{nieuwenhuis_cremers_pami12}. Just like
the authors, we adapt the variance of the spatial kernel $\rho_i$ to
the distance of the current pixel $x$ from the nearest user scribble
of this class: $\rho_i(x) = \alpha |x - x_{v_i}|_2$ where $x_{v_i}$ is
the closest scribble location of all pixels in segment $i$ and
$\alpha$ a scaling factor, which we set to $1.3$. The idea is that
each color kernel is weighted by a spatial kernel in order to account
for the uncertainty in the estimator at locations far away from the
scribble points and for the certainty very close to them. 

We will now extend the spatial color variation of \cite{nieuwenhuis_cremers_pami12} to a spatially varying texture distribution. To that end, we need to derive the probability that a patch $\mathbf{s}_x$ belongs to segment $i$, \ie $\mathcal{P}(\mathbf{s}_x,x| l(x) = i)$. As our goal is to extract local textural information in the vicinity of a pixel $x$, we multiply each patch element-wise with a Gaussian mask to assign more weight to the central pixels prior to normalization according to Section \ref{sec:textural_similarity}. From these patches, we compute the approximated co-support of a textural representative of each set of scribble points according to Equation \eqref{eq:approx_cosupp}, \ie
\begin{equation} 
\mathbf{c}_i = \med(\{\iota_\sigma(\mathbf{Os}_{x_{ij}})\}_{j=1}^{m_i}).
\end{equation}
Using this, we assign to each pixel $x$ the a posteriori probability of belonging to class $i$ depending on the corresponding patch as
\begin{align}\label{eq:post_text}
  \mathcal{P}(\mathbf{s}_x,x|\l\! =\! i) = \frac{\exp(-\tfrac{1}{\beta_i}\|\mathbf{c}_i-\iota_\sigma(\mathbf{O}\mathbf{s}_x)\|_1)}{\sum_{j=1}^{n} \exp(-\frac{1}{\beta_j}\|\mathbf{c}_j-\iota_\sigma(\mathbf{O}\mathbf{s}_x)\|_1)},
\end{align}
where the parameters $\beta_i>0$ control the variance of $l$. It can be interpreted as a measure of how well we trust the distance for deciding to which class $x$ belongs. %
Large values of $\beta_i$ assign a pixel to each of the classes with approximately equal probability, whereas small values of $\beta_i$ assign $x$ to the most similar class with very high probability. By setting $\beta_i$ proportional to $\rho_i$ we obtain a spatially varying texture distribution, which favors spatially close texture classes.

\paragraph{Unsupervised Segmentation}
The approach introduced above can also be used for unsupervised image segmentation. In this case no color, texture, or location samples are given a priori. Location likelihoods cannot be obtained without sample data or user input, so we rely on color and texture only. To find segment classes, we first apply the k-means on the color pixels to obtain color classes and the k-medians on the approximated image patches' co-supports to obtain texture classes. With that, the initial segments are given as the combinations of all texture classes with all color classes (because the same texture with different color should be treated as a different segment). The segment class representatives are given as the centroids of the corresponding texture cluster and color cluster. With these representatives, we assign each pixel a texture- and a color-likelihood as in \eqref{eq:post_text}. The number of segments is automatically reduced by a minimum description length (MDL) prior, which is added to the variational approach introduced in the next section.

\subsection{Variational Formulation}
Based on the segment probabilities $\PP\big(I(x), \mathbf{s}_x, x\big|\, l\! =\! i\big)$ given in \eqref{eq:joined}, we now define an energy optimization problem for the tasks of supervised and unsupervised segmentation. To this end, we specify the prior $\mathcal{P}(l)$ in \eqref{eq:bayes2} to favor segmentation regions of shorter boundary
\begin{equation}
  \label{eq:prior}
  \PP(l) \propto \exp\big(-\tfrac{\lambda}{2}\sum_{i=1}^n
    \mbox{Per}_g(\Omega_i) \big), 
\end{equation}
where $\mbox{Per}_g(\Omega_i)$ denotes the perimeter of each region
$\Omega_i$ measured in metric $g: \Omega \to \mathbb{R}^+$. For
example, the commonly used choice
\begin{equation*}
g(x) = \tfrac{1}{2 \gamma} \exp\left(- \tfrac{|\nabla I(x)|}{\gamma}\right), \;\;\; \gamma = \tfrac{1}{|\Omega|} \int_{\Omega} | \nabla I(x) | \; dx,
\end{equation*}
favors boundaries coinciding with strong intensity gradients
$|\nabla I(x)|$ and, thus, prevents oversmoothed boundaries.  We set
$\gamma = 5$. The weighting parameter $\lambda \in
[0,\infty]$ balances the impact of the data term $f_i$ and the
boundary length.

For supervised
segmentation, the number of segments is known, but for unsupervised
segmentation the number of segments must be determined automatically by means of an MDL prior~\cite{yuan_boykov10}. 
Let us define the function $\delta_i:
\Omega \to \{0,1\}$,\begin{equation}
\delta_i(\Omega) = 
\begin{cases}
1, & \Omega_i \neq \emptyset \\
0, & \text{otherwise,}
\end{cases}
\end{equation} 
which indicates if the label $i$ occurs in the segmentation. By multiplying \eqref{eq:prior} with the MDL prior
\begin{equation} \label{eq:mdl}
\PP(l) \propto \exp \big(-\nu \sum_{i=1}^n\delta_i(\Omega) \big),
\end{equation}
we penalize the occurrence of a label by a cost $\nu > 0$. With increasing $\nu$, the number of labels in the segmentation result is reduced and the number of segments automatically optimized. For supervised segmentation $\nu$ can be set to zero.

Instead of maximizing the a posteriori distribution \eqref{eq:apost}, we minimize its negative logarithm, \ie the energy
\begin{align}\label{eq:energy1}
\E=
\sum_{i=1}^n\tfrac{\lambda}{2}\mbox{Per}_g(\Omega_i) + \nu
\delta_i(\Omega) + \!\int\limits_{\Omega_i}\! f_i(x) \; dx,
\end{align}
where $f_i(x) \!=\! -\log \PP\big(I(x),\mathbf{s}_x,x\big|\,l(x)\! =\! i\big)$ in \eqref{eq:joined} in combination with \eqref{eq:color} and \eqref{eq:post_text}.

\section{Minimization via Convex Relaxation}
\label{sec:optimization}
Problem \eqref{eq:energy1} is the
continuous equivalent to the Potts model, whose solution is known to be
NP-hard. However, a computationally tractable convex relaxation of this functional has been proposed in \cite{chambolle-et-al08,Chan-et-al-06,lellmann08,Pock-et-al-iccv09,zach_et_al08}.
Due to the convexity of the problem the resulting solutions have the
following properties: Firstly, the segmentation is independent of the
initialization. Secondly, we obtain globally optimal segmentations for
the case of two regions and near-optimal -- in practice often globally
optimal -- solutions for the multi-region case.  In addition, the
algorithm can be parallelized leading to average computation times of
$6$ seconds per image on standard GPUs.


\subsection{Conversion to a Convex Differentiable Problem}
To apply convex relaxation techniques, we first represent the $n$
regions $\Omega_i$ by the indicator function $u \in \mbox{BV}(\Omega,\{0,1\})^n$, where
\begin{equation}
  \label{eq:thetas}
 	u_i(x) =
	\begin{cases}
		1, \quad\text{if } x \in \Omega_i \\
		0, \quad\text{otherwise}
	\end{cases}\quad i \in \{1,..,n\}.
\end{equation}
Here $\mbox{BV}$ denotes the functions of bounded variation, \ie
functions with a finite total variation.  
For a valid segmentation we require that  the sum of all indicator functions at each location $x \in \Omega$ amounts to one, so each pixel
is assigned to exactly one label. Hence, 
\begin{equation}
\BB= \big\{u \in \text{BV}(\Omega, \{0,1\})^n\;\Big|\;\sum_{i=1}^n u_i = 1\big\}.
\end{equation}
denotes the set of valid segmentations. 
To rewrite energy \eqref{eq:energy1} in terms of the
indicator functions $u_i$, we have to rewrite the label prior $\PP(l)$.  \\
We will start with the boundary length prior in \eqref{eq:prior}. The
boundary of the set indicated by $u_i$ can be written by means of the
total variation.  Let $D\,u_i$ denote the distributional derivative of
$u_i$ (which is $D\,u_i = \nabla u_i \; dx$ for differentiable $u_i$),
$\xi_i \in C^1_c(\Omega, \mathbb{R}^2)$ the dual variables with
$C^1_c$ the space of smooth functions with compact support.

Then, following the coarea formula \cite{federer96} the weighted perimeter of $\Omega_i$ is equivalent to the weighted total variation
\begin{eqnarray}
 	\frac{\lambda}{2} \mbox{Per}_g(\Omega_i) &=& \frac{\lambda}{2}\int_\Omega  \; |D \,u_i \;| = \sup_{\xi_i \in \KK_g} \int_{\Omega} \xi_i \; D\, u_i \; \nonumber\\
 	&=& \sup_{\xi_i \in \KK_g} -\sum_{i=1}^n\int_{\Omega} u_i \; \mbox{div} \; \xi_i \; dx  \label{intbyparts}
\end{eqnarray}
$\mbox{with} \;\; \KK_g\! =\! \left\{\xi \in C^1_c(\Omega, \mathbb{R}^2)
	\Big| \: |\xi(x)| \leq \tfrac{\lambda g(x)}{2}, x \in \Omega \right\}$ \cite{zach_et_al08}.
The latter transformation \eqref{intbyparts} follows from integration by parts and the compact support of the dual variables $\xi_i$.

Following \cite{yuan_boykov10}, let us rewrite the minimum description length prior in \eqref{eq:mdl} which is given by $\nu \sum_{i=1}^n \delta_i(\Omega)$ in terms of the region indicator functions $u_i$. To this end, we define maximum value variables \mbox{$m_i \in \{0,1\}, \;\;m_i := \max_{x \in \Omega} u_i(x)$} which yields
\begin{equation}
\nu \sum_{i=1}^n \delta_i(\Omega) = \nu \sum_{i=1}^n \max_{x\in\Omega}u_i(x) = \nu \sum_{i=1}^n m_i.
\end{equation}
To obtain a convex optimization problem 
we relax the set $\BB$
\begin{equation}
\tilde{\BB} = \big\{u \in \text{BV}(\Omega, [0,1])^n \;\Big|\;\sum_{i = 1}^n u_i = 1\big\}.
\end{equation}
Furthermore, the constraints on the maximum value variables $m_i$ are relaxed to the convex constraints 
$
m_i \geq u_i(x) 
$
for all $x \in \Omega$, and the $m_i$ are minimized.
These constraints are introduced into the optimization problem using Lagrangian multipliers $\mu \in \mathcal{M}$, where $\mathcal{M}:= \left\{\mu:\Omega \to [-\infty, 0]^n\right\}$. With this, we finally obtain the convex optimization problem
\begin{eqnarray}
   \min_{\stackrel{u \in \tilde{\BB}}{m \in [0,1]^n}} \sup_{\stackrel{\xi_i \in \KK_g}{\mu \in \mathcal{M}}} \!\!\!\!\!\!\!\!&&
   \sum_{i=1}^n  \int_{\Omega} u_i \,f_i \; dx - \int_{\Omega} u_i \mbox{ div }
     \xi_i \; dx + \nonumber\\
    && \nu \, m_i + \int_{\Omega}\mu_i(x) (m_i - u_i(x)) \; dx. 
\label{eq:sets}
\end{eqnarray}

\subsection{Implementation}

To solve the relaxed convex optimization problem, we employ a primal
dual-algorithm proposed in~\cite{Pock-et-al-iccv09}. Essentially, it
consists of alternating a projected gradient descent in the primal
variables $u_i$ and $m_i$ with projected gradient ascent in the dual
variables $\xi_i$ and $\mu_i$. An over-relaxation step in the primal
variables gives rise to auxiliary variables $\bar{u_i}$ and
$\bar{m_i}$: 

\vspace{-0.18cm}
\begin{eqnarray} \label{algo}
	\xi_i^{t+1} &=& \Pi_{\KK_g} \Big(\xi_i^t + \tau_\xi \nabla \bar{u}_i^t\Big) \\[1mm] 
        \mu_i^{t+1} &=& \Pi_{\mathcal{M}} \Big(\mu_i^t + \tau_\mu (\bar{m}_i^t - \bar{u}_i^t)\Big) \nonumber\\[1mm]
        m_i^{t+1} &=& \Pi_{[0,1]}\Big(m_i^t - \tau_m (\nu + \int_\Omega \mu(x) \; dx) \Big) \nonumber \\[1mm]
	u_i^{t+1} &=& \Pi_{\tilde{\BB}} \Big(u_i^t - \tau_u (-\mbox{ div } \xi_i^{t+1} + f_i - \mu_i^{t+1})\Big) \nonumber\\[1mm] 
 	\bar{u}_i^{t+1} &=&  u_i^{t+1} + (u_i^{t+1} - u_i^t) = 2 u_i^{t+1} - u_i^t \nonumber\\[1mm]
 	\bar{m}_i^{t+1} &=&  m_i^{t+1} + (m_i^{t+1} - m_i^t) = 2 m_i^{t+1} - m_i^t \nonumber
\end{eqnarray}
where $\Pi$ denotes the projections onto the respective convex sets and the $\tau$ denote step sizes for primal and dual variables. These are optimized based on~\cite{pock_chambolle11}.
The projections onto $\KK_g, \mathcal{M}$ and $[0,1]^n$ are straightforward, the projection onto the
simplex $\tilde{\BB}$ is given in \cite{michelot86}. As shown in \cite{Pock-et-al-iccv09}, the algorithm \eqref{algo} provably converges to a minimizer of the relaxed problem.

Due to the relaxation we may end up with non-binary solutions
$u_i \in \bar{\BB}$. To obtain binary solutions in the set $\BB$, we assign each
pixel to the label with maximum value $u_i$, \ie $l(x) = \argmax_i
u_i(x)$.  This operation is known to preserve optimality in case of
two regions \cite{Chan-et-al-06}. In the multi-region case optimality
bounds can be computed from the energy difference between the
minimizer of the relaxed problem and its reprojected
version. Typically the projected solution deviates less than 1\% from
the optimal energy, \ie the results are very close to global
optimality \cite{nieuwenhuis_cremers_pami12}. 

\section{Experiments and Results}
\label{sec:experiments}
To evaluate the proposed algorithm we apply it to the interactive Graz benchmark \cite{santner10} for supervised segmentation and to images from the
well-known Berkeley segmentation database \cite{martin_et_al01} for unsupervised segmentation. We compare our approach against state-of-the-art segmentation algorithms. For all experiments we use a patch size of $9 \times 9$, and a two times over complete analysis operator, \ie $k=2*81$. We do not require any training of the operator on the training set but use it as is avoiding over fitting to specific benchmarks. The parameter $\sigma$ in \eqref{eq:map_tsm} required to measure the textural similarity was set to $\sigma = 0.01$. 
The textural similarity analysis is based only on highly parallelizable filter operations. Due to the inherently parallel structure of the optimization problem in \eqref{algo}, the algorithm can be efficiently implemented on graphics hardware. The experiments were carried out on an Intel Core i7-3770 3.4 GHz CPU with an NVIDIA Geforce GTX 580 GPU. 

The average computation time per image of the Berkeley database with sizes of 321x481 was six seconds. For comparison, the following runtimes per image were reported: for Yang et al. \cite{yang_et_al08} 3 minutes, for Mobahi et al. \cite{mobahi_et_al11} 1 minute, for Santner et al. \cite{santner10} 2 seconds and for Nieuwenhuis et al. \cite{nieuwenhuis_cremers_pami12} 1.5 seconds.

\subsection{Results on the Graz Benchmark} 
We employ the Graz benchmark \cite{santner10} for evaluating our supervised segmentation algorithm. It consists of 262 scribble-ground truth pairs from 158 natural images containing between 2 and 13 user labeled segments. Here, we used a brush size of 13 pixels in diameter for scribbling as done by Santner et al. \cite{santner10}, and set $\lambda = 2000$. To rank our method, we compare our results with the Random Walker algorithm by Grady \cite{Grady-06}, a learning based approach that combines color and texture features by Santner \cite{santner10}, and the approach based on spatially varying color distribution by Nieuwenhuis and Cremers \cite{nieuwenhuis_cremers_pami12}, which combines color and spatial information. Table \ref{tab:graz_benchmark} shows the average Dice-score \cite{dice45} for all methods. This score compares the overlap of each region $\Omega_i$ with its ground truth $\bar{\Omega}_i$
\begin{equation}\label{eq:dice}
dice(\Omega_1,.. \Omega_n) = \frac{1}{n} \sum_{i=1}^n \frac{2|\bar{\Omega}_i \cap \Omega_i|}{|\bar{\Omega}_i| + |\Omega_i|}.
\end{equation}
The results show that the proposed method outperforms state-of-the-art supervised segmentation approaches.
Figure \ref{fig:graz} shows qualitative comparisons for a some images of the Graz benchmark, where texture is important to separate the regions. The sign on the wall can only be distinguished from the background by texture, the ground beneath the walking men changes color due to lighting and can only be recognized by texture, and the towels on the table are also segmented correctly only by using texture information. The average computation time on the Graz Benchmark is 2 seconds, which is along the lines of Santner et al. \cite{santner10} with 2 seconds and Nieuwenhuis et al. \cite{nieuwenhuis_cremers_pami12} with 1.5 seconds.

%

\begin{table}
\setlength{\tabcolsep}{2mm}
\begin{center} 
\begin{tabular}{|l|c|} \hline 
\rr Method & Score \\ \hline
\rr Grady~\cite{Grady-06}, Random Walker &  0.855  \\
\rr Santner et al.~\cite{santner10}, RGB, no texture & 0.877  \\
\rr Nieuwenhuis \& Cremers~\cite{nieuwenhuis_cremers_pami12}, space-constant & 0.889 \\
\rr Santner~\cite{santner10}, CIELab plus texture  & 0.927\\
\rr Nieuwenhuis \& Cremers~\cite{nieuwenhuis_cremers_pami12}, space-varying  & 0.931 \\
\rr \textbf{Proposed} (spatially varying co-sparse texture) & \textbf{0.935} \\  \hline
\end{tabular}\vspace{0.2cm}
\caption{Comparison of the average Dice-score \eqref{eq:dice} to
  state-of-the-art supervised segmentation approaches by Grady
  \cite{Grady-06}, Santner et al. \cite{santner10} (RGB color based,
  CIELab color and texture) and Nieuwenhuis and
  Cremers (space-constant and space-varying color model)
  \cite{nieuwenhuis_cremers_pami12} on the Graz benchmark.}
\vspace{-0.6cm}
\label{tab:graz_benchmark} 
\end{center}  
\end{table} 


\begin{figure*}[tp]
  \setlength{\tabcolsep}{1mm}
  \begin{center}
    \begin{tabular}{ccccc}
      \includegraphics[width=0.18\linewidth]{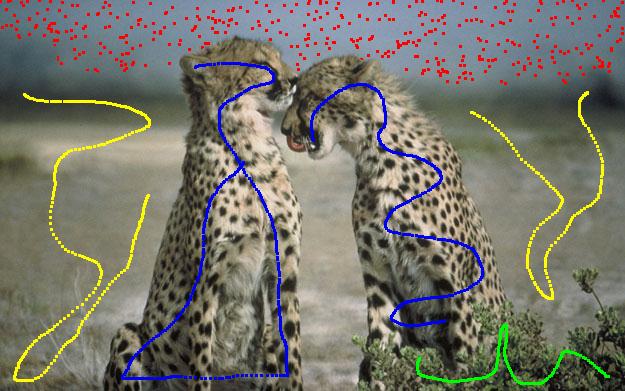}&
      \includegraphics[width=0.18\linewidth]{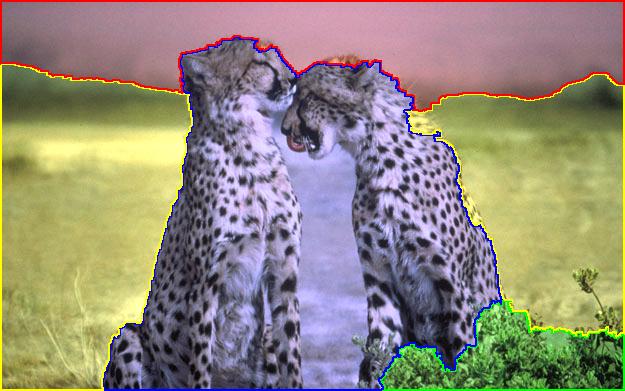}&
       \includegraphics[width=0.18\linewidth]{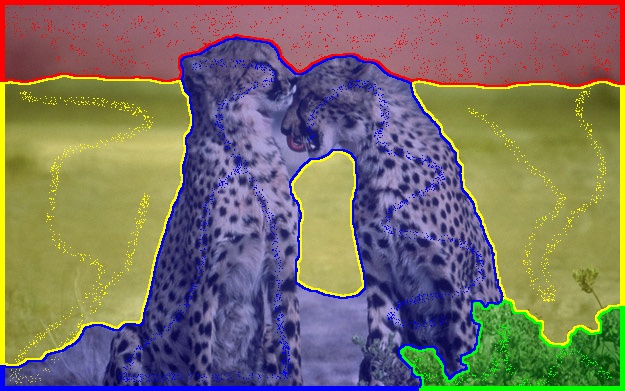}&
      \includegraphics[width=0.18\linewidth]{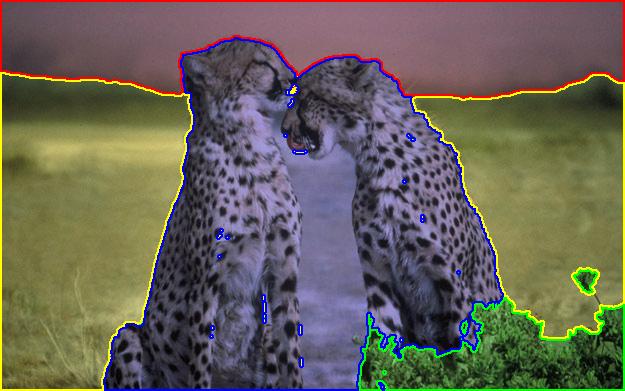}&
      \includegraphics[width=0.18\linewidth]{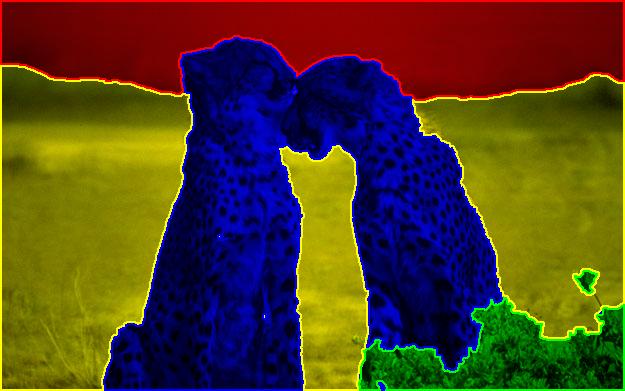} \\
      \includegraphics[width=0.18\linewidth]{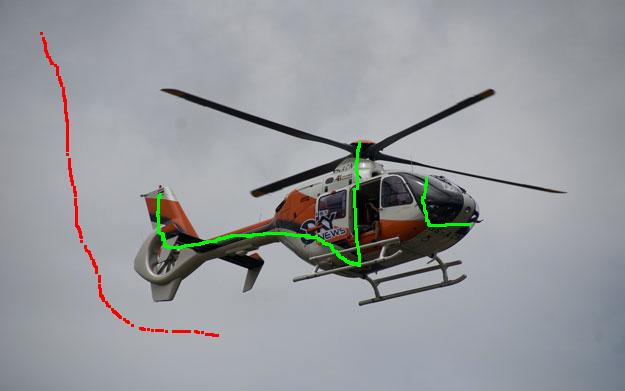}&
      \includegraphics[width=0.18\linewidth]{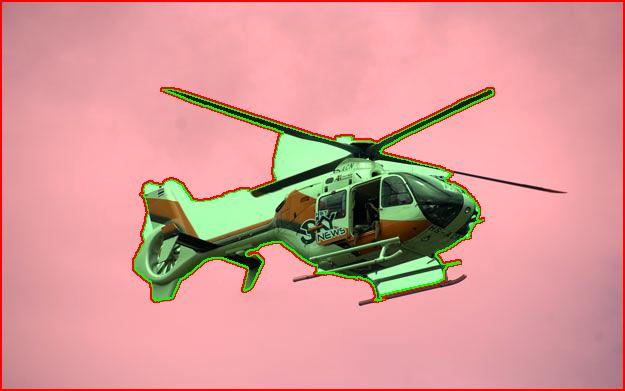}&
       \includegraphics[width=0.18\linewidth]{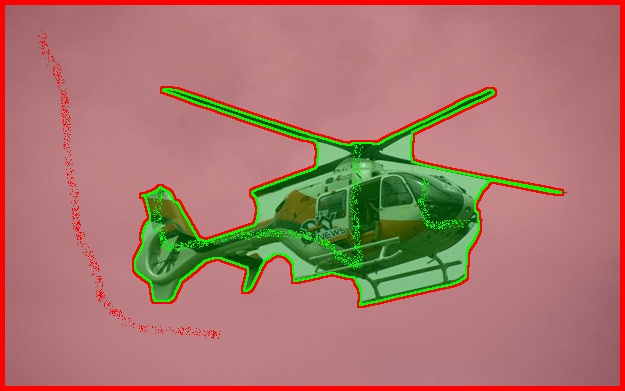}&
      \includegraphics[width=0.18\linewidth]{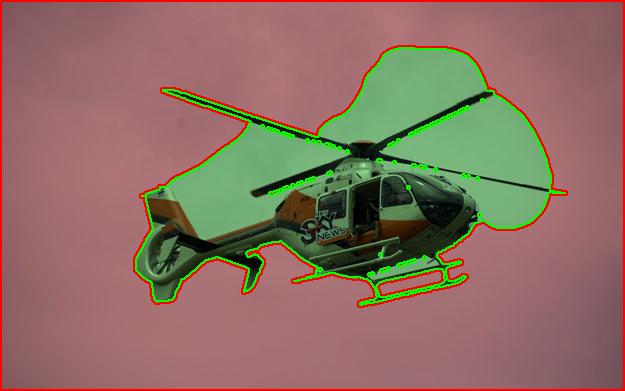}&
      \includegraphics[width=0.18\linewidth]{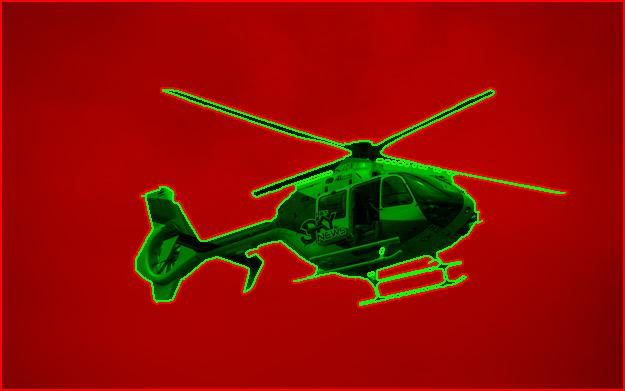} \\
      \includegraphics[width=0.18\linewidth]{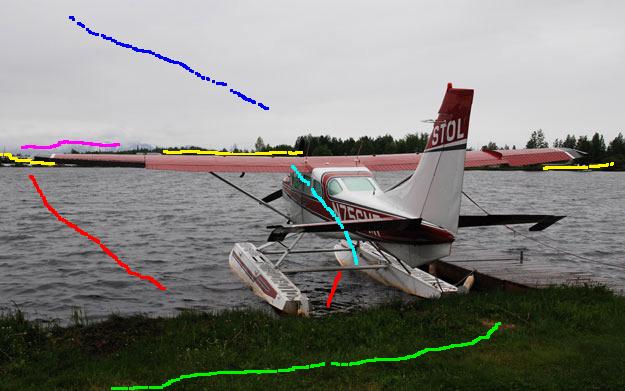}&
      \includegraphics[width=0.18\linewidth]{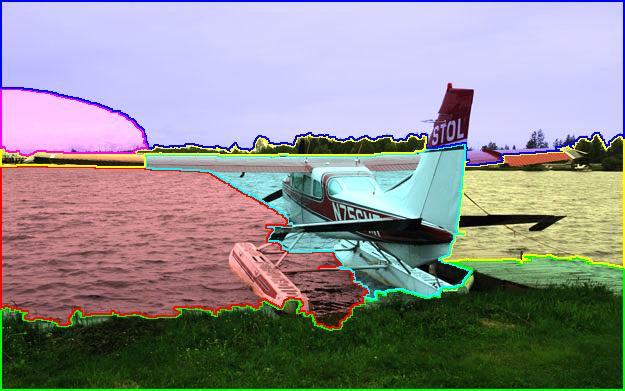}&
       \includegraphics[width=0.18\linewidth]{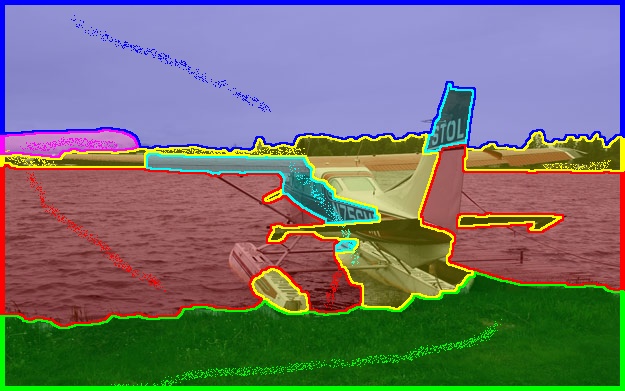}&
      \includegraphics[width=0.18\linewidth]{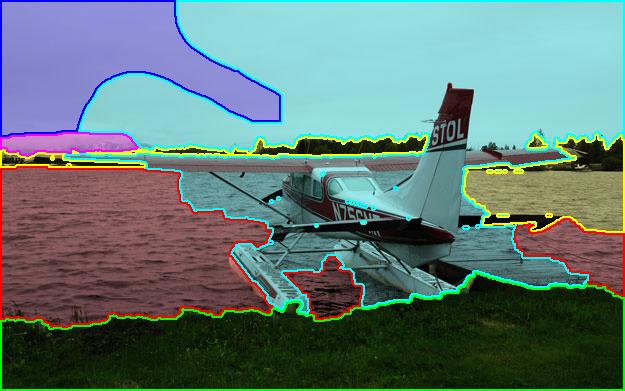}&
      \includegraphics[width=0.18\linewidth]{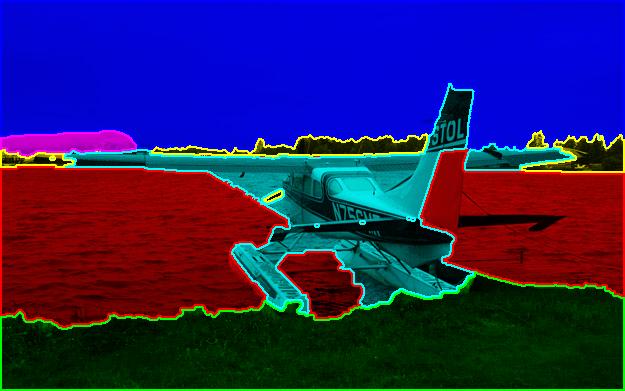} \\
      \includegraphics[width=0.18\linewidth]{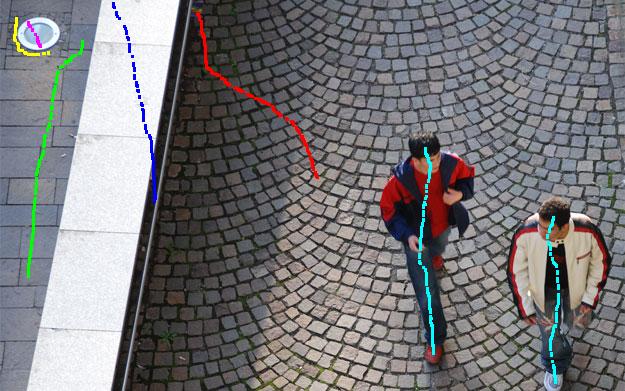}&
      \includegraphics[width=0.18\linewidth]{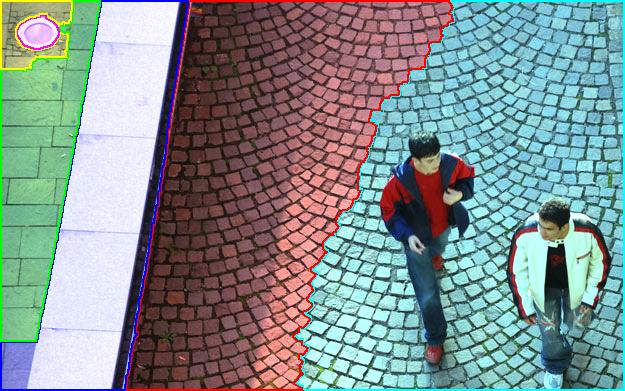}&
       \includegraphics[width=0.18\linewidth]{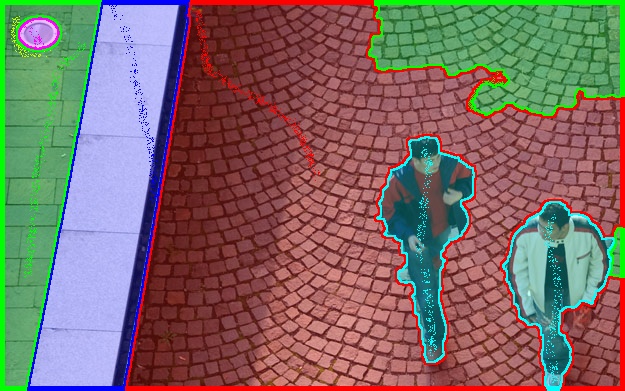}&
      \includegraphics[width=0.18\linewidth]{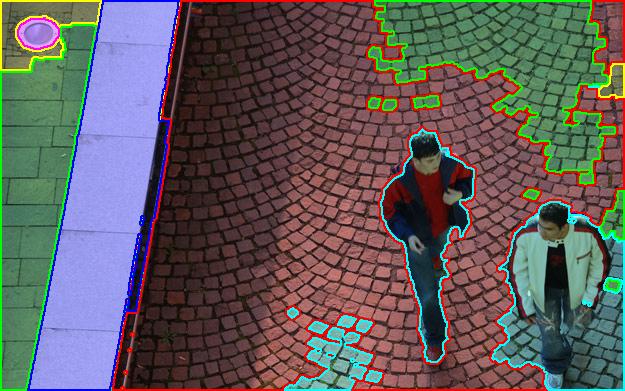}&
      \includegraphics[width=0.18\linewidth]{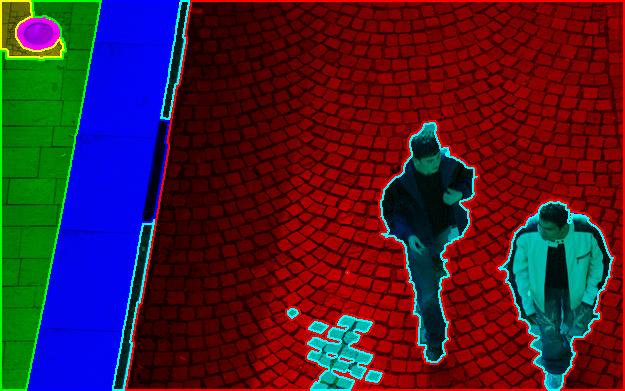} \\
      \includegraphics[width=0.18\linewidth]{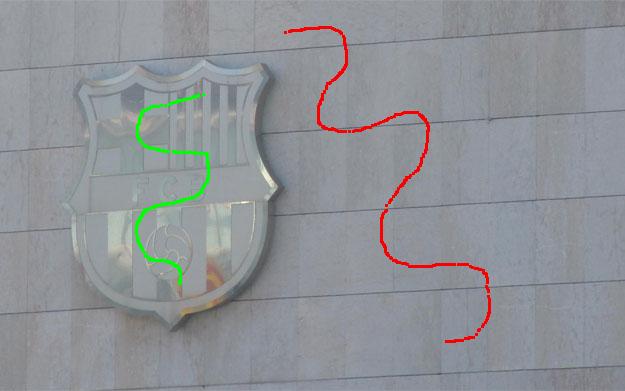}&
      \includegraphics[width=0.18\linewidth]{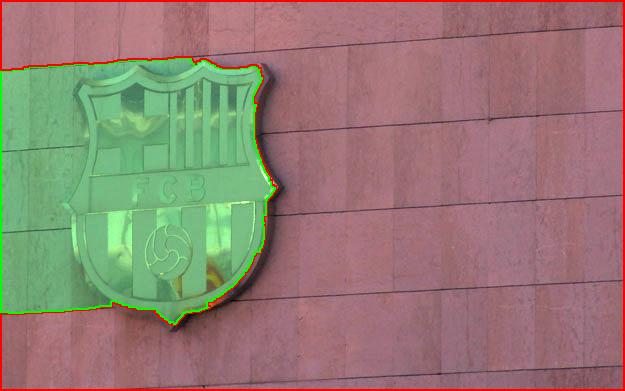}&
       \includegraphics[width=0.18\linewidth]{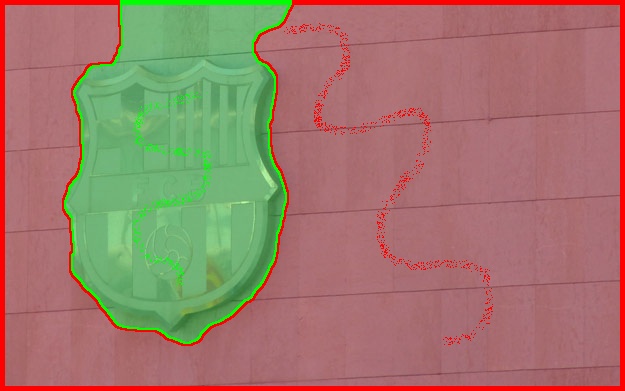}&
      \includegraphics[width=0.18\linewidth]{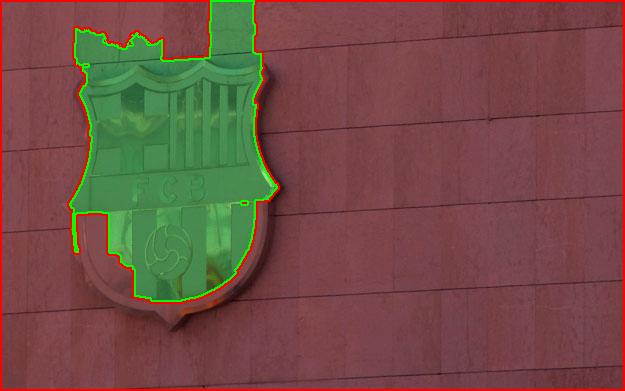}&
      \includegraphics[width=0.18\linewidth]{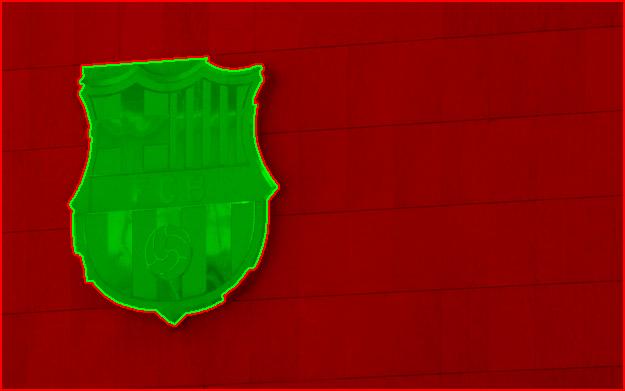} \\
      \includegraphics[width=0.18\linewidth]{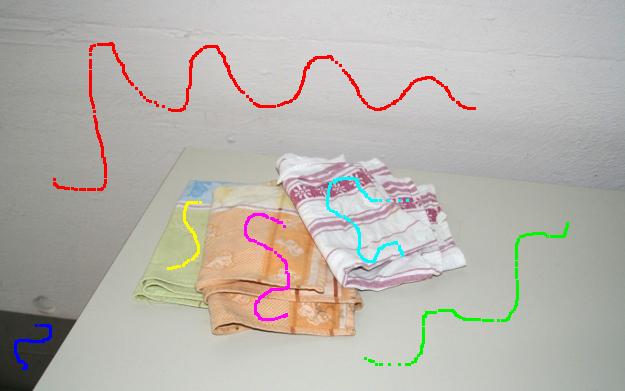}&
      \includegraphics[width=0.18\linewidth]{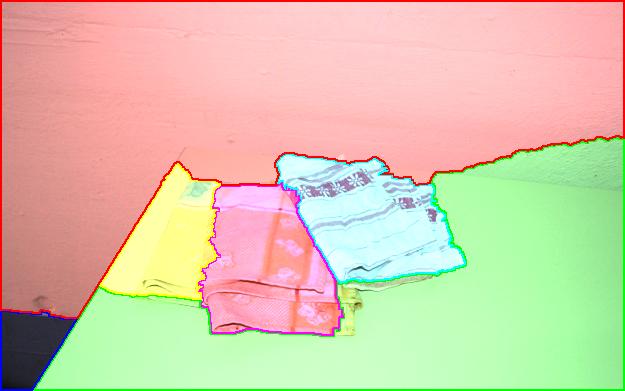}&
       \includegraphics[width=0.18\linewidth]{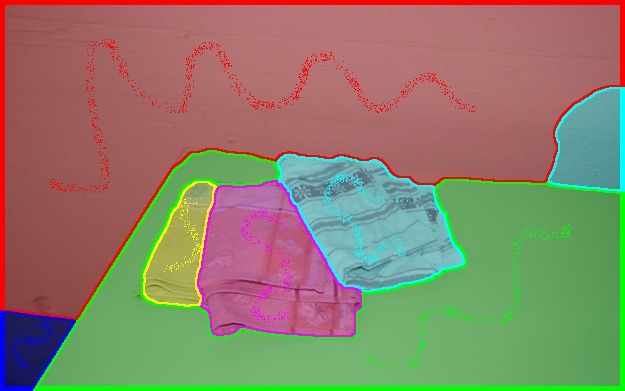}&
      \includegraphics[width=0.18\linewidth]{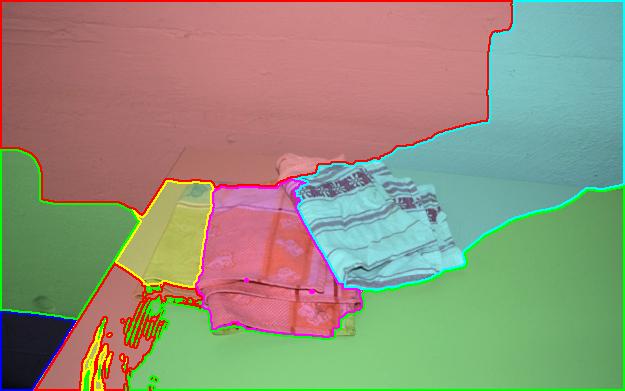}&
      \includegraphics[width=0.18\linewidth]{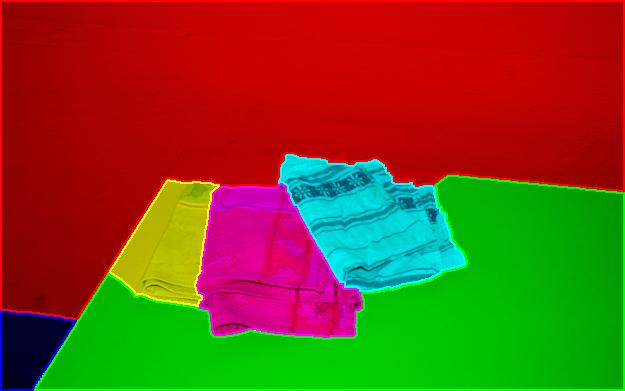} \\
      \includegraphics[width=0.18\linewidth]{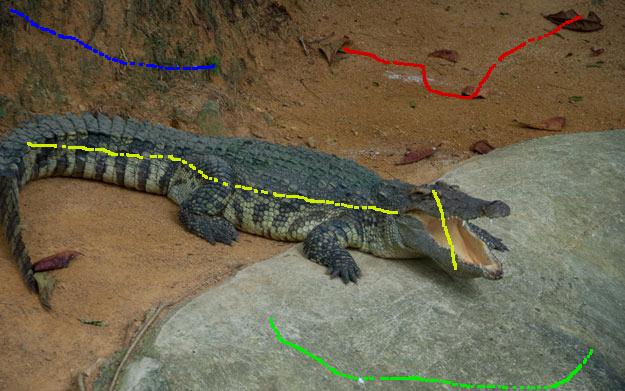}&
      \includegraphics[width=0.18\linewidth]{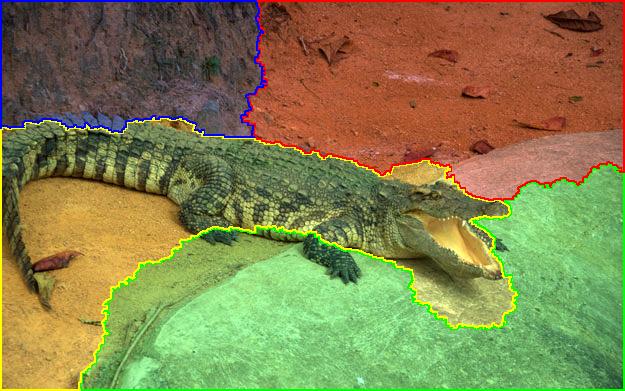}&
       \includegraphics[width=0.18\linewidth]{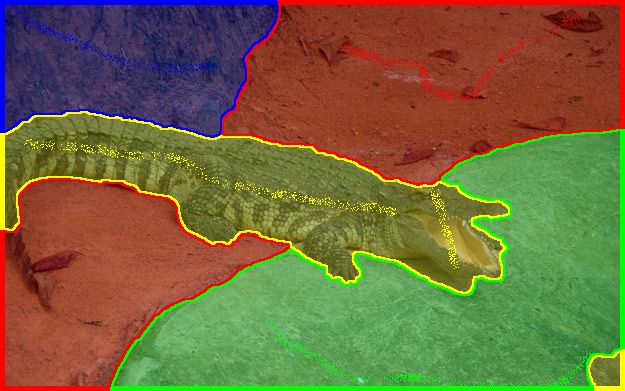}&
      \includegraphics[width=0.18\linewidth]{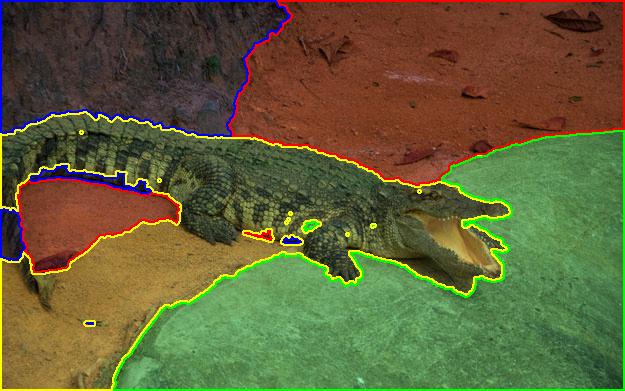}&
      \includegraphics[width=0.18\linewidth]{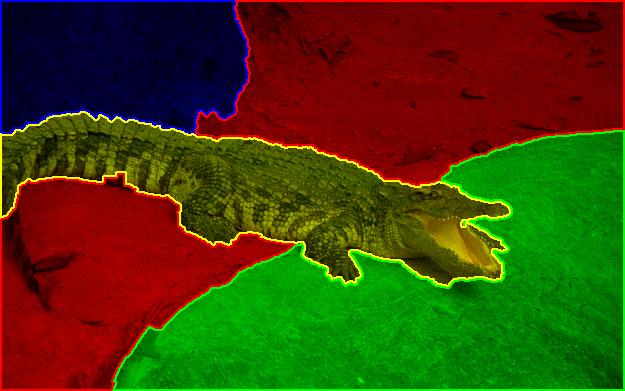} \\
      a) Original & b) Grady \cite{Grady-06} & c) Santner et
      al. \cite{santner10} & \parbox{0.18\linewidth}{\centering d)
        Nieuwenhuis and Cremers \cite{nieuwenhuis_cremers_pami12}} &
      e) Proposed \vspace{-0.2cm} 
    \end{tabular} 
  \end{center}
  \caption{Comparison of supervised segmentation to the approaches by
    Grady \cite{Grady-06}, Santner et al. \cite{santner10} and
    Nieuwenhuis and Cremers \cite{nieuwenhuis_cremers_pami12} on the
    Graz interactive segmentation benchmark.} \vspace{-0.4cm}
  \label{fig:graz}
\end{figure*}
   
\subsection{Results on the Berkeley Segmentation Database}
The Berkeley segmentation database \cite{martin_et_al01}
contains 300 natural images, 200 for training and 100 for testing. It
serves for the evaluation of contour detectors and unsupervised
segmentation algorithms. The ground truth is given by the overlay of
hand-drawn human segmentations. We set $\lambda\! =\! 6$, $\nu\! =\! 1100$, and the initial number of regions to $n\! =\! 16$.
Benchmark results with respect to boundary displacement error
(BE), probabilistic rand index (PRI), global consistency error (GCE)
and variation of information (VOI) are given in Table
\ref{tab:berkeley}. While the performance of the proposed method
according to the similarity to human annotations is only mediocre, we
observed that in many cases the segmentations are superior to
those of state-of-the-art methods.

Figure \ref{fig:bsds} shows comparisons of computed segmentations with
the state-of-the-art segmentation methods, which do not compute edges but closed object segments. We compare against the
methods by Yang et al. \cite{yang_et_al08},
Mignotte~\cite{mignotte12} and Mobahi et al.~\cite{mobahi_et_al11}. 
The results show that the proposed method visually outperforms the
other algorithms on a number of images where it better captures the
objects in the scene. The resulting segments are more coherent (see
for example the first image of the brown animal) and the
segment boundaries follow more closely the actual object boundaries
instead of texture edges (see for example the legs of the wolf, which
are preserved, or the skiing person). We show more results in the supplemental material.

The average computation time per image of our algorithm on the Berkeley database with sizes of 321x481 was six seconds, which is due to the larger number of labels for this benchmark. This is a strong improvement in efficiency compared to other methods for unsupervised segmentation, \eg Yang et al. \cite{yang_et_al08} reported 3 minutes and Mobahi et al. \cite{mobahi_et_al11} 1 minute per image.


\begin{table}
\setlength{\tabcolsep}{2mm}
\begin{center} 
\begin{tabular}{|l|c|c|c|c|} \hline 
\rr Method & BE & PRI & VOI & GCE \\ \hline
\rr NCut~\cite{Shi-Malik-97} & 17.15&  0.7242 & 2.9061 & 0.2232   \\
\rr FH~\cite{Felzenszwalb-et-al-04} &  16.67 & 0.7139 & 3.3949 & \textbf{0.1746}   \\
\rr Meanshift~\cite{Comaniciu-meer-02} & 14.41 & 0.7958 & 1.9725 & 0.1888  \\
\rr \textbf{Proposed} & 13.66 & 0.7244 & 2.6156 & 0.3412  \\ 
\rr Mobahi et al.\cite{mobahi_et_al11}$\!$ & 12.681 & \textbf{0.807} & \textbf{1.705} & -  \\ 
\rr Yang et al.~\cite{yang_et_al08} & 9.8962 & 0.7627 & 2.0236 & 0.1877  \\  
\rr Mignotte~\cite{mignotte12}  & \textbf{8.9951} & 0.7882 & 2.3035 & 0.2114   \\  \hline
\end{tabular}\vspace{0.2cm}
\caption{Comparison of the proposed unsupervised segmentation method to state-of-the-art approaches on the Berkeley benchmark. Even though we do not outperform current approaches we obtain a competitively low boundary displacement error. Qualitative results in Figure \ref{fig:bsds} show that our results essentially concentrate on the objects in the image instead of marking various texture edges.} \vspace{-0.4cm}
\label{tab:berkeley} 
\end{center}
\end{table}

      \begin{figure*}[tp]
          \setlength{\tabcolsep}{1mm}
        \begin{center}
          \begin{tabular}{ccccc}

             \includegraphics[width=0.18\linewidth]{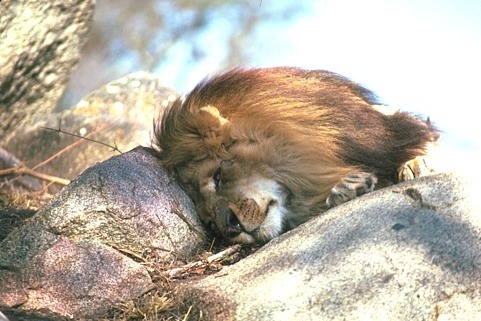}&
             \includegraphics[width=0.18\linewidth]{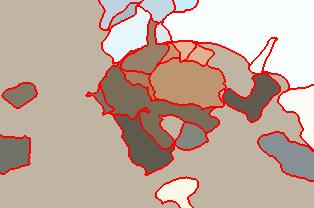}&
             \includegraphics[width=0.18\linewidth]{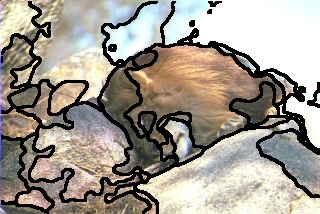} &
             \includegraphics[width=0.18\linewidth]{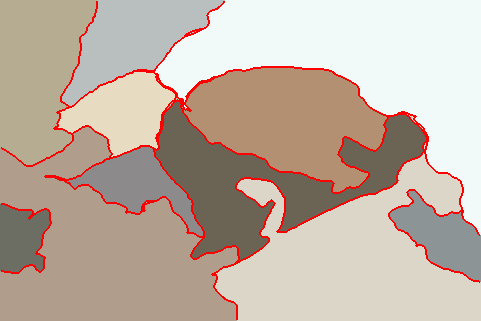} &
             \includegraphics[width=0.18\linewidth]{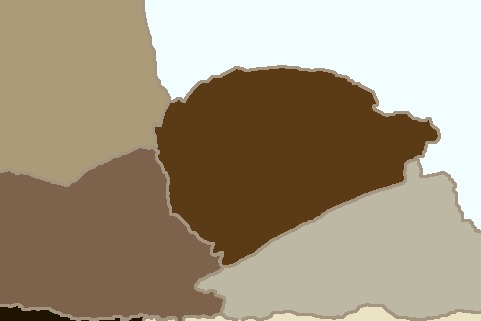} \\

             \includegraphics[width=0.18\linewidth]{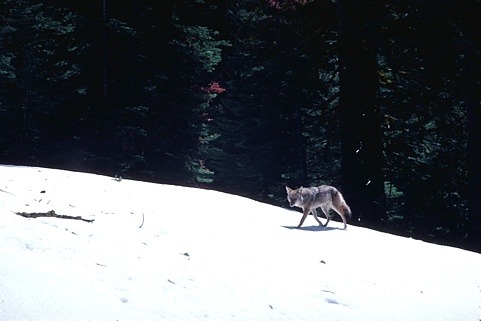}&
             \includegraphics[width=0.18\linewidth]{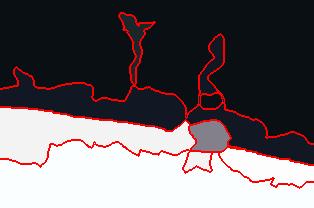}&
             \includegraphics[width=0.18\linewidth]{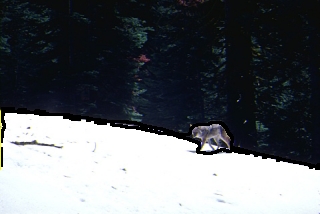} &
             \includegraphics[width=0.18\linewidth]{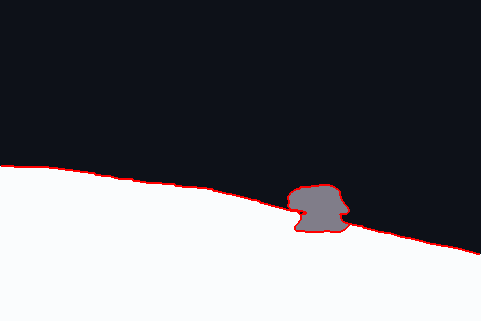} &
             \includegraphics[width=0.18\linewidth]{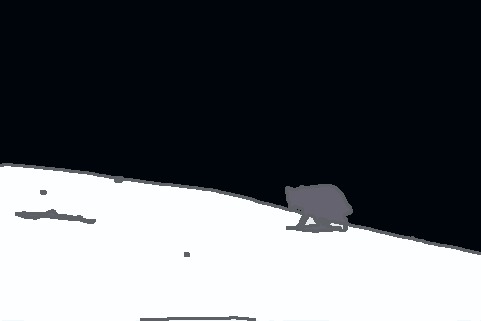} \\

             \includegraphics[width=0.18\linewidth]{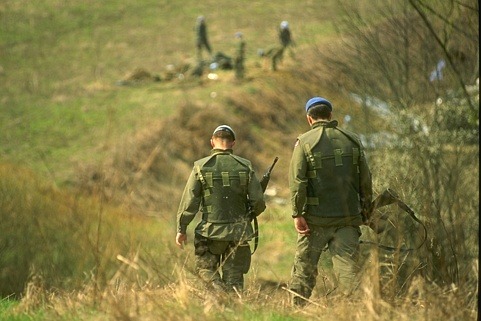}&
             \includegraphics[width=0.18\linewidth]{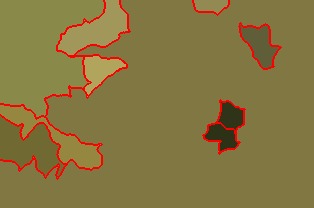}&
             \includegraphics[width=0.18\linewidth]{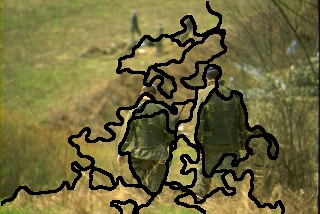} &
             \includegraphics[width=0.18\linewidth]{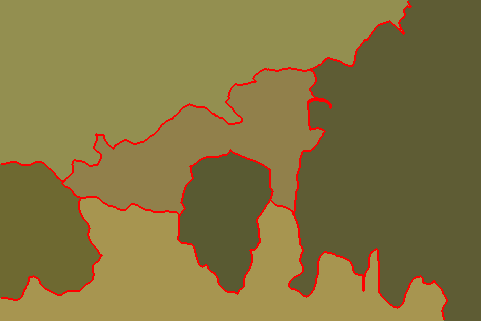} &
             \includegraphics[width=0.18\linewidth]{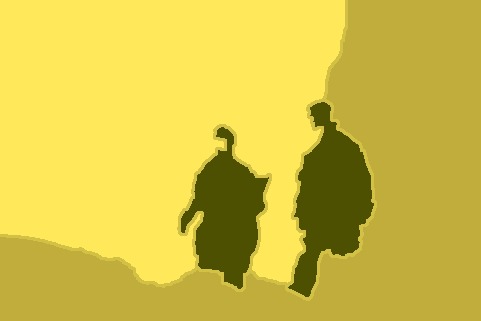} \\

             \includegraphics[width=0.18\linewidth]{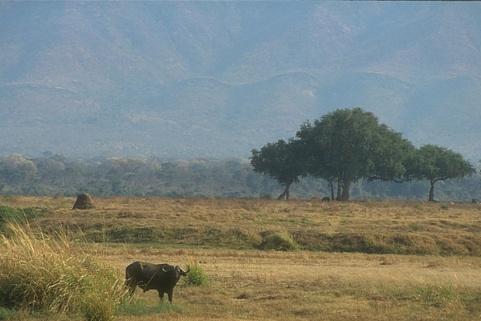}&
             \includegraphics[width=0.18\linewidth]{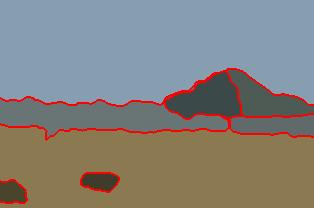}&
             \includegraphics[width=0.18\linewidth]{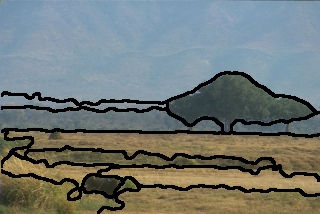} &
             \includegraphics[width=0.18\linewidth]{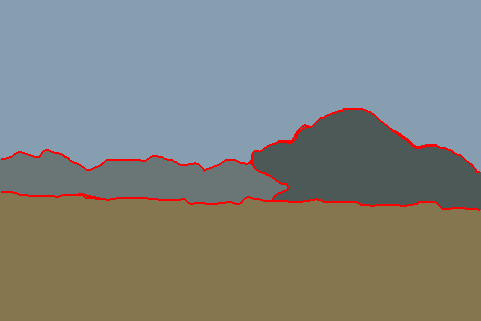} &
             \includegraphics[width=0.18\linewidth]{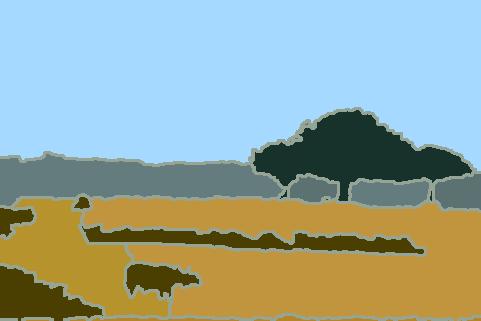} \\

             \includegraphics[width=0.18\linewidth]{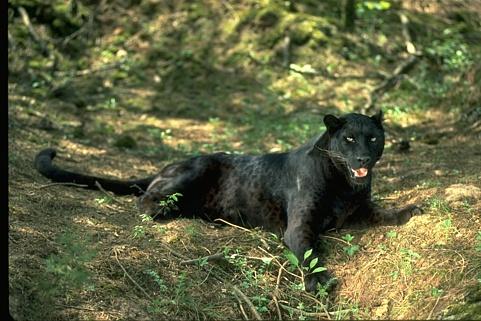}&
             \includegraphics[width=0.18\linewidth]{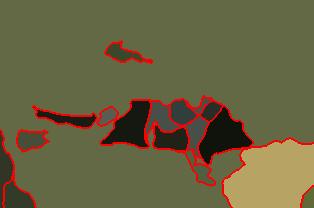}&
             \includegraphics[width=0.18\linewidth]{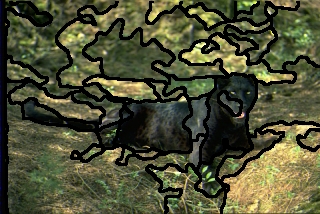} &
             \includegraphics[width=0.18\linewidth]{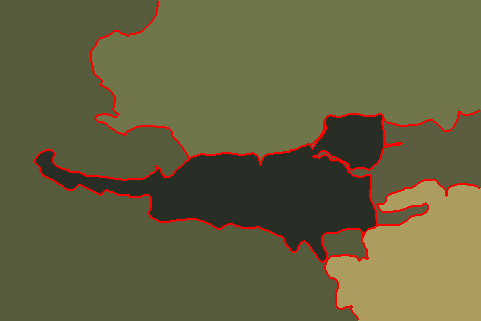} &
             \includegraphics[width=0.18\linewidth]{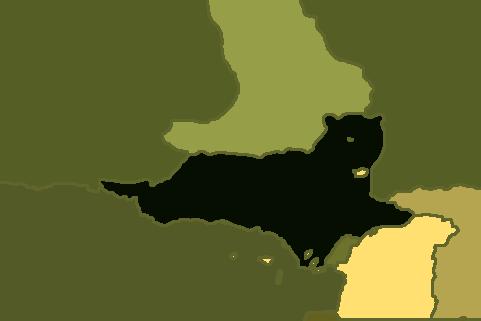} \\

              \includegraphics[width=0.18\linewidth]{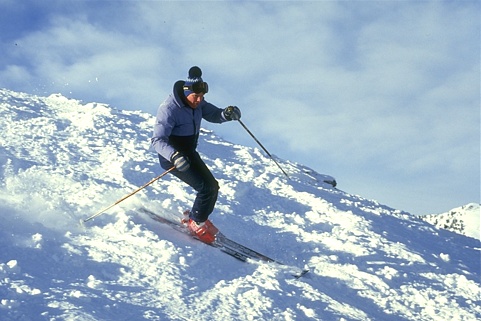}&
              \includegraphics[width=0.18\linewidth]{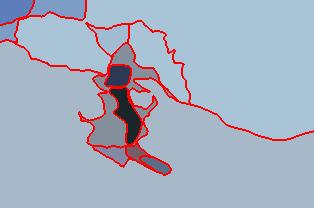}&
              \includegraphics[width=0.18\linewidth]{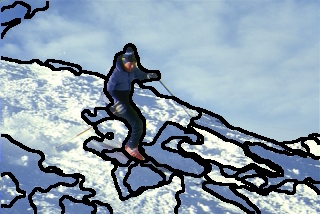} &
              \includegraphics[width=0.18\linewidth]{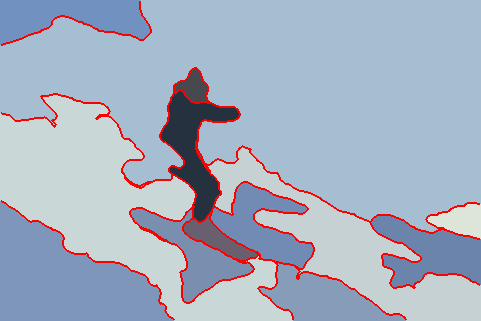} &
              \includegraphics[width=0.18\linewidth]{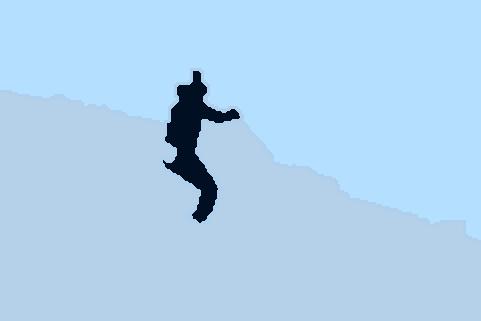} \\

            Original & Yang et al.~\cite{yang_et_al08} & Mignotte~\cite{mignotte12} & Mobahi et al.~\cite{mobahi_et_al11} & Proposed approach \\
          \end{tabular}
        \end{center}
        \caption{Comparison of the proposed approach to state-of-the-art methods on the Berkeley Segmentation Database.} \vspace{-0.4cm}
\label{fig:bsds}
        
      \end{figure*}

\section{Conclusion}

We introduced a framework for segmentation of natural images which is applicable to both, supervised and unsupervised image segmentation.
A new measure of textural similarity which is based on the co-sparse representation of image patches
 is proposed. From this measure, a data likelihood is derived and integrated in a Bayesian maximum a posteriori estimation scheme in order to combine color, texture, and location information. The arising cost functional is minimized by means of convex relaxation techniques. With our efficient GPU implementation of the convex relaxation, the overall algorithm for multiregion segmentation converges within about two to six seconds for typical images. Moreover, the approach outperforms state-of-the-art methods on the Graz segmentation benchmark.




{\small
\bibliographystyle{ieee}
\bibliography{allrefs}
}

\end{document}